\documentclass[10pt,a4paper]{article}

\usepackage[a4paper,top=2.54cm,bottom=2.54cm,left=1.91cm,right=1.91cm,
            headheight=24pt,headsep=8pt,footskip=20pt]{geometry}
\usepackage[T1]{fontenc}
\usepackage{newtxtext,newtxmath}          
\usepackage{graphicx}
\usepackage{booktabs}
\usepackage{amsmath}
\usepackage{array}
\usepackage{enumitem}
\usepackage[font=small,labelfont=bf,labelsep=period,justification=centering, skip=4pt]{caption}
\usepackage{titlesec}
\usepackage{fancyhdr}
\usepackage[hidelinks]{hyperref}

\renewcommand{\thesection}{\arabic{section}.}
\renewcommand{\thesubsection}{\arabic{section}.\arabic{subsection}}
\titleformat{\section}{\fontsize{11}{13}\selectfont\bfseries}{\thesection}{0.4em}{}
\titleformat{\subsection}{\fontsize{10}{12}\selectfont\bfseries}{\thesubsection}{0.5em}{}
\titlespacing*{\section}{0pt}{8pt}{3pt}
\titlespacing*{\subsection}{0pt}{6pt}{2pt}
\setlist{nosep,leftmargin=1.4em}

\begin{document}

\begin{center}
{\fontsize{14}{17}\selectfont\bfseries
Land--Water Co-Simulation of an Amphibious Quadruped Leg: Joint Loads, Speed Effects and Hydrodynamic Sensitivity\par}
\vspace{8pt}
{\fontsize{12}{14}\selectfont Akash Kumar\textsuperscript{1,*},
Jagadeesh Kadiyam\textsuperscript{2}\par}
\vspace{4pt}
{\footnotesize
\textsuperscript{1}Centre for AI and Robotics, Indian Institute of Technology Mandi, Himachal Pradesh, India\\
\textsuperscript{2}Department of Ocean Engineering and Naval Architecture, Indian Institute of Technology Kharagpur, West Bengal, India\\
\textsuperscript{*}Corresponding author: \texttt{d23147@students.iitmandi.ac.in}\par}
\end{center}

\noindent\textbf{Abstract}\par
\noindent Legged robots that walk on the seabed and cross the shoreline must work with joint actuators sized for both media, yet the loads a leg meets in water and on land are rarely compared under identical conditions. This paper presents a simulation framework in which the same three-joint quadruped leg, with the same leg trajectory and joint controller, is run on land in Gazebo and in still water in Stonefish, with buoyancy, pressure drag, skin friction and added mass applied to every link. Because the leg is almost neutrally buoyant, water removes the gravity load that dominates on land: at 0.20 m/s the RMS knee torque falls from 0.26 to 0.08 N\,m while the joint motion is unchanged. The water loads, however, are dynamic and increase with speed. In a sweep from 0.10 to 0.30 m/s, the RMS hip-flexion torque in water rises from 0.06 to 0.41 N\,m and exceeds the land value above 0.23 m/s; above 0.25 m/s, walking in water requires 40\% more joint work than on land at 0.30 m/s. The peak torque, in contrast, remains set by land at every speed. The added-mass force is the largest hydrodynamic force on the shank at all speeds. The simulated water torques lie between two Morison strip-theory estimates and follow their shape and speed scaling, and a sensitivity study shows that the added mass controls the result, not the drag coefficient. The findings in this research paper show that actuators sized for land suffice in water up to about 0.23 m/s, above which their continuous rating must be increased, and identify the added mass of the thin-walled links as the next quantity to measure in the planned hardware tests.

\vspace{3pt}

\section{Introduction}

Quadruped robots have become capable machines on land. Platforms such as ANYmal~\cite{hutter2016} and MIT Cheetah~3~\cite{bledt2018} climb stairs, cross rubble and recover from slips, because legs need only isolated footholds and can step over obstacles that stop wheels and tracks. These robots, however, are designed for dry ground: their actuators, sealing, sensors and controllers assume that the legs move through air. At most, they tolerate rain and splashes; they cannot wade through deep water, walk on a submerged bottom, or continue a mission after the ground gives way to a river, a flooded street, or the sea.

Many important tasks lie exactly at this boundary between land and water. In flood-hit areas, rescuers must search streets, buildings and fields covered by muddy water full of debris, where boats cannot pass, and aerial vehicles cannot see below the surface~\cite{murphy2014}. Coastal and military operations need reconnaissance across beaches and the surf zone, where a
vehicle must leave the water, cross the shore and return. Bridge piers, dams, harbour walls and pipeline landfalls must be inspected both above and below the waterline, and shallow seabeds, wetlands and reefs must be monitored without disturbing the sediment or the habitat. An Amphibious Quadruped Robot (AQR), which walks on land, enters the water, and continues on the bottom with the same body and legs, could perform all of the above tasks without changing platforms. Underwater, legs keep their advantages: a legged robot can stand still on the bottom against a current, carry tools close to a structure and walk without stirring up the sediment~\cite{picardi2020,jun2013}.

Several of the existing robots have demonstrated parts of this capability. The amphibious robot AQUA walks on land and swims with its legs~\cite{dudek2007}, a salamander robot switches between walking and swimming~\cite{ijspeert2007}, and robots that change their limb shape have moved between land and water by adapting their morphology~\cite{baines2022}. On the seabed, the legged robots SILVER~\cite{picardi2020} and CR200~\cite{jun2013} have shown that walking underwater is feasible and efficient, and models of underwater legged locomotion have highlighted how strongly buoyancy and hydrodynamic forces change the dynamics of a leg~\cite{calisti2017}. What these studies do not provide is a direct, controlled comparison of the loads that the same leg, with the same leg trajectory and controller, meets in the two media such as water and land.

Such a comparison is necessary because an amphibious quadruped has a single set of actuators for both media. For its designer, a practical question follows: are actuators chosen for walking on land also adequate underwater, and up to which walking speed? On land, a leg mainly carries its own weight and the ground reaction force; underwater, buoyancy cancels most of the weight, but pressure drag, skin friction, and added mass oppose every motion and grow with speed. Which of these effects dominates depends on the leg's geometry, its density and its gait, and cannot be read from either environment alone.

Open-source simulators make such a comparison possible. Gazebo~\cite{koenig2004} is widely used for legged robots on land, and Stonefish~\cite{cieslak2019} and the UUV Simulator~\cite{manhaes2016} model the hydrodynamics of marine
robots. In this work, we use them side by side, with the same robot description, leg trajectory, and controller, so that any differences in joint loads are due to the medium. The main contributions of this paper are:

\begin{itemize}
  \item A land-water co-simulation of one leg of an amphibious quadruped, in which only the environment changes between the two simulations;
  \item A hydrodynamic model in Stonefish whose drag, skin-friction and added-mass terms follow the standard formulations;
  \item A speed sweep that identifies the walking speed at which the water loads exceed the land loads; and
  \item An analytic (Morison strip-theory) check of the simulated water loads and a sensitivity study that ranks the model parameters.
\end{itemize}

Section~2 describes the leg, controller and simulation set-up. Sections~3 and~4 compare land and water at one speed and across speeds, respectively. Section~5 examines how far these results can be trusted, and Section~6 summarises the findings and discusses their implications for the design of the leg. Section~7 outlines future work.

\section{Leg Model and Simulation Framework}

\subsection{Leg and test stand}

\begin{figure}[htbp]
  \centering
  \includegraphics[width=0.62\linewidth]{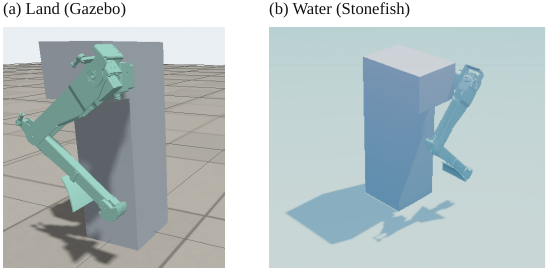}
  \caption{The leg on its test stand in the two simulations: (a) on land in Gazebo and (b) submerged in a still-water tank in Stonefish. In both simulations, the same robot description, leg trajectory and joint controller are used.}
  \label{fig:setup}
\end{figure}

The study uses the front-right (FR) leg of an amphibious quadruped robot as shown in Figure~\ref{fig:setup}. It has three revolute joints: hip abduction/adduction (HAA), hip flexion/extension (HFE) and knee flexion/extension (KFE). Masses, inertias, and joint limits are taken from the CAD-derived robot description (Table~\ref{tab:params}); each joint is limited to a torque of 6\,N\,m and an angular velocity of 2\,rad/s. The thigh (0.273\,kg, 270\,cm$^3$) and the shank (0.183\,kg, 181\,cm$^3$) have a mean density of 1.01\,g/cm$^3$, so the leg is almost neutrally buoyant in fresh water. The leg is bolted to a rigid pillar, and in water, the hip is 0.17\,m below the free surface.

\begin{table}[htbp]
  \centering
  \caption{Leg dimensions, leg trajectory and simulation parameters.}
  \label{tab:params}
  \small
  \begin{tabular}{@{}llll@{}}
    \toprule
    Parameter & Value & Parameter & Value \\
    \midrule
    Thigh / shank length & 0.280 m / 0.302 m & Leg trajectory & 1.0 s cycle (nominal) \\
    HAA / thigh / shank mass & 0.175 / 0.273 / 0.183 kg & Stride / foot lift & 0.10 / 0.04 m \\
    Thigh / shank volume & 270 / 181 cm$^3$ & Swing fraction & 0.5 \\
    Joint limits & 6 N m, 2 rad/s & Nominal speed & 0.20 m/s \\
    PD gains HAA, HFE & $K_p$ = 150 N m/rad, $K_d$ = 3 N m s/rad & Physics step & 1 ms (both simulators) \\
    PD gains KFE & $K_p$ = 80 N m/rad, $K_d$ = 1.5 N m s/rad & Water & fresh, still, 1000 kg/m$^3$ \\
    \bottomrule
  \end{tabular}
\end{table}

\subsection{Leg trajectory and joint control}

The leg follows a trajectory in which the foot sweeps backward in a straight line for half of the cycle and returns along a swing path for the other half. The swing path is a cycloid in the horizontal direction, with zero speed relative to the body at lift-off and touch-down. The forward speed is the stride divided by the stance time, $v = s/(T/2)$. Inverse kinematics converts the foot path into joint references, which are sent at 250\,Hz to a joint-space PD controller running in the 1\,kHz physics loop:

\begin{equation}
  \tau_j = K_{p,j}\,(q_{\mathrm{ref},j} - q_j) + K_{d,j}\,(\dot q_{\mathrm{ref},j} - \dot q_j),
  \label{eq:pd}
\end{equation}

where $q_j$ is the angle of joint $j$, $q_{ref,j}$ is the reference angle of joint $j$, $K_{p,j}$ is proportional gain for the joint $j$, $K_{d,j}$ is derivational gain for the joint $j$ and $\tau_j$ is torque to the joint $j$. On land, this law runs in the \texttt{ros2\_control} trajectory controller~\cite{chitta2017}; in water, the same law, with the same gains and reference interpolation, runs inside the Stonefish physics step. Both simulations are driven by the same ROS~2 nodes~\cite{macenski2022}, and joint angles, velocities and torques are recorded.

\subsection{Land and water environments}

The land simulation uses Gazebo Harmonic with a 1\,ms step. The water simulation uses Stonefish~1.6~\cite{cieslak2019} with a 1\,ms step: the leg stands in a 20\,$\times$\,20\,m glass-walled tank of still fresh water, without waves or currents. Stonefish computes the water loads on the triangulated surface of each link at every step. We checked its default per-element formulas against the standard drag laws and found that the pressure-drag and skin-friction terms were not dimensionally consistent forces, and that the added mass, computed from an ellipsoid enclosing each mesh, was about ten times the displaced water for these thin links. These terms were therefore replaced by the formulations below, as options in a patch to Stonefish.

\subsection{Hydrodynamic model}

For each link, with water density $\rho$ and gravity $g$, the buoyancy acts at the centre of the displaced volume $V$, and each surface element of area $A$ that meets the flow contributes pressure drag and skin friction:

\begin{equation}
  \mathbf F_b = \rho g V\,\mathbf e_z,\qquad
  d\mathbf F_d = -\tfrac12 \rho C_d\,A_p\,|\mathbf v|\,\mathbf v,\qquad
  d\mathbf F_f = -\tfrac12 \rho C_f\,A\,|\mathbf v_t|\,\mathbf v_t ,
  \label{eq:forces}
\end{equation}

where $\mathbf v$ is the element velocity relative to the water, $A_p = A\,(-\mathbf v\cdot\mathbf n)/|\mathbf v|$ the area projected normal to the flow, $\mathbf v_t$ the tangential velocity and $C_d \le 1$ a shape-dependent coefficient. The skin-friction coefficient follows the flat-plate laws~\cite{schlichting2017}, $C_f = 1.328/\sqrt{Re}$ (laminar) or $0.074/Re^{0.2}$ (turbulent, $Re > 5\times10^5$), with $Re = |\mathbf v_t|L/\nu$ and $L$ the link length. The added mass $m_a$ of a link is computed from an ellipsoid with the proportions of the link and the volume $V$, and is added to the link inertia; the resulting force is

\begin{equation}
  \mathbf F_a = -m_a\,\dot{\mathbf v}_G .
  \label{eq:addedmass}
\end{equation}

Where $F_A$ is the added-mass force, $m_a$ is the added mass, and $V_G$ is the velocity of the link's center of gravity. For the leg, $\rho g V$ is 2.65\,N (thigh) and 1.77\,N (shank), and $m_a$ is 0.201 and 0.133\,kg.

\subsection{Test cases and metrics}

The nominal case is a walk at 0.20\,m/s. For the speed sweep, the stride length is kept at 0.10\,m and the leg trajectory cycle time is shortened from 2.0 to 0.67\,s, giving a speed of the value 0.10, 0.15, 0.20, 0.25 and 0.30\,m/s with the same foot path; at 0.30\,m/s, the HFE joint reaches 1.85\,rad/s, close to its 2\,rad/s limit. For the sensitivity study, the drag coefficient $C_d$ and the added mass $m_a$ are each scaled by factors of 0, 0.5, 1.5, and 2 at 0.20 and 0.30\,m/s. 
Besides the RMS and peak joint torques, the positive joint work per cycle is calculated using the following equation:

\begin{equation}
  W = \frac{1}{N}\sum_j \int_{\text{N cycles}} \max\!\left(\tau_j\dot q_j,\,0\right)dt ,
  \label{eq:work}
\end{equation}

Equation~\ref {eq:work} measures the energy the actuators must deliver. Where $N$ is the number of cycles and $\tau_j\dot q_j$ is the mechanical power of the joint $j$. 

\section{Land versus Water at Nominal Speed}

\begin{figure}[htbp]
  \centering
  \includegraphics[width=\linewidth]{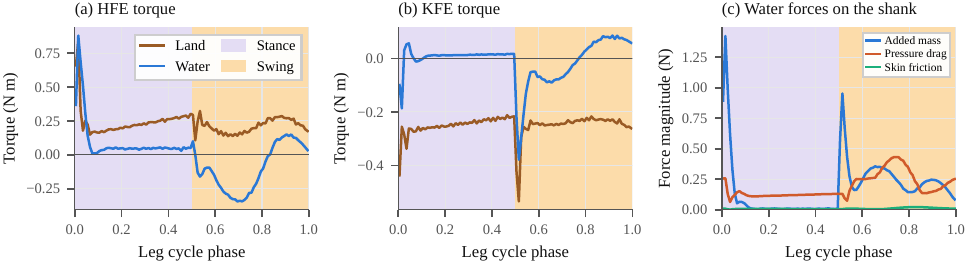}
  \caption{Walking at 0.20\,m/s, averaged over the steady leg trajectory cycles: (a) HFE and (b) KFE torque on land and in water; (c) magnitude of the water forces on the shank. The shaded half of each panel is the swing phase.}
  \label{fig:nominal}
\end{figure}

At 0.20\,m/s the leg moves in the same way in both media. The joint-tracking error is 0.03$^\circ$ RMS on land and 0.07$^\circ$ in water, and the foot paths differ by only 0.8\,mm RMS. The torques needed for this motion, however, differ strongly (Figure~\ref{fig:nominal}(a), and \ref{fig:nominal}(b)).

On land, the torques are dominated by gravity. The mean HFE torque is 0.23\,N\,m and the mean KFE torque is $-0.25$\,N\,m throughout the cycle, and the HAA joint carries a constant 0.26\,N\,m to hold the leg's weight sideways. In water, buoyancy almost exactly balances the weight, and these mean values fall to below 0.01\,N\,m. From land to water,the RMS torque drops from 0.26 to 0.04\,N\,m at HAA, from 0.25 to 0.20\,N\,m at HFE and from 0.26 to 0.08\,N\,m at KFE.

What remains in water is a dynamic load: it arises only from the motion of the leg (drag, added mass and inertia) and would vanish if the leg stood still. The HFE torque in water is small (about 0.05\,N\,m) during stance and swings between $-0.3$ and $+0.15$\,N\,m during the swing, when the leg moves fastest, with short peaks at lift-off and touch-down. Figure~\ref{fig:nominal}(c) shows where these loads come from. On the shank, the added-mass force (0.30\,N RMS) and the pressure drag (0.21\,N RMS) are of the same order, while the skin friction is twenty times smaller (0.011\,N). The added-mass force peaks at lift-off and touch-down: there the cycloid swing path, which starts and ends at zero speed relative to the body, meets the stance motion at 0.2\,m/s, and the foot velocity changes abruptly. The thigh, which moves less, sees much smaller forces (0.07, 0.02 and 0.003\,N). At this speed, i.e., 0.2\,m/s, water reduces the joint loads. Since drag and added mass grow with speed while the gravity load does not, the next question is how long this advantage lasts.

\section{Effect of Walking Speed}

\begin{figure}[htbp]
  \centering
  \includegraphics[width=\linewidth]{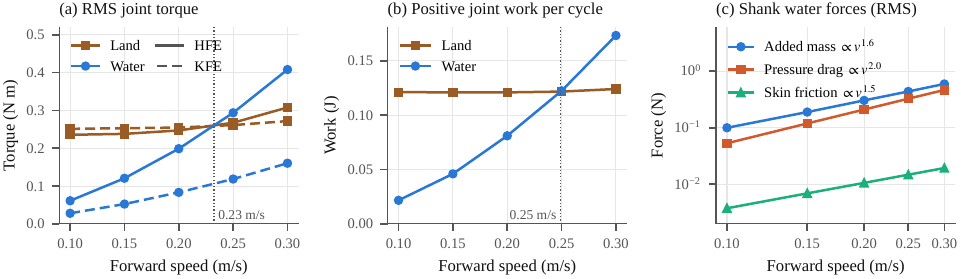}
  \caption{Speed sweep with a fixed 0.10\,m stride: (a) RMS joint torque for land and water, (b) positive joint work per leg trajectory cycle for land and water; (c) RMS water forces on the shank (logarithmic axes) with the fitted speed exponents. Dotted lines mark the speeds where water overtakes land.}
  \label{fig:speed}
\end{figure}

Figure~\ref{fig:speed} summarises the speed sweep. On land, the RMS torques change little with speed (HFE from 0.24 to 0.31\,N\,m) and the positive joint work stays at 0.12\,J per cycle, because lifting the leg against gravity costs the same at every speed. In water, both grow steeply: from 0.10 to 0.30\,m/s the RMS HFE torque rises from 0.06 to 0.41\,N\,m and the work per cycle from 0.022 to 0.173\,J.

Figure~\ref{fig:speed}(a) and \ref{fig:speed}(b) conclude that the RMS HFE torque in water exceeds the land value above 0.23\,m/s, and the joint work above 0.25\,m/s. At 0.30\,m/s, the leg needs 33\% more RMS HFE torque and 40\% more work per cycle in water than on land. The KFE torque stays below its land value over the whole range (0.16 versus 0.27\,N\,m at 0.30\,m/s), and the peak torques on land are equal to or higher than in water at every speed (HFE 1.77 versus 1.51\,N\,m at 0.30\,m/s), because on land the inertial peaks add to the gravity load.

The speed exponents of the shank forces (Figure~\ref{fig:speed}(c)) agree with the force laws. The pressure drag grows as $v^{2.0}$, as equation~\ref{eq:forces} predicts for a motion whose shape does not change with speed, and the skin friction as $v^{1.5}$, the laminar flat-plate scaling ($Re < 5\times10^5$). The added-mass force grows only as $v^{1.6}$, although a scaled motion would give $v^2$; the reason is the velocity steps at lift-off and touch-down, whose duration is set by the joint servo rather than by the leg trajectory. Added mass remains the largest water force at every tested speed, but drag grows faster (exponent of 2.0 versus 1.6), so the gap narrows. At 0.30 m/s, they are 0.60 N (added mass) and 0.47 N (drag). Extending the trends, drag would overtake added mass at about 0.55 m/s. That is beyond the reach of this leg due to the joint speed limit.

These results rely on the simulator's hydrodynamic model. Before they are used in design, the next section checks whether the simulated water loads are realistic and identifies the model parameter that mostly affects them.

\section{Reliability of the Simulated Water Loads}

\subsection{Analytic check at the joints}

\begin{figure}[htbp]
  \centering
  \includegraphics[width=\linewidth]{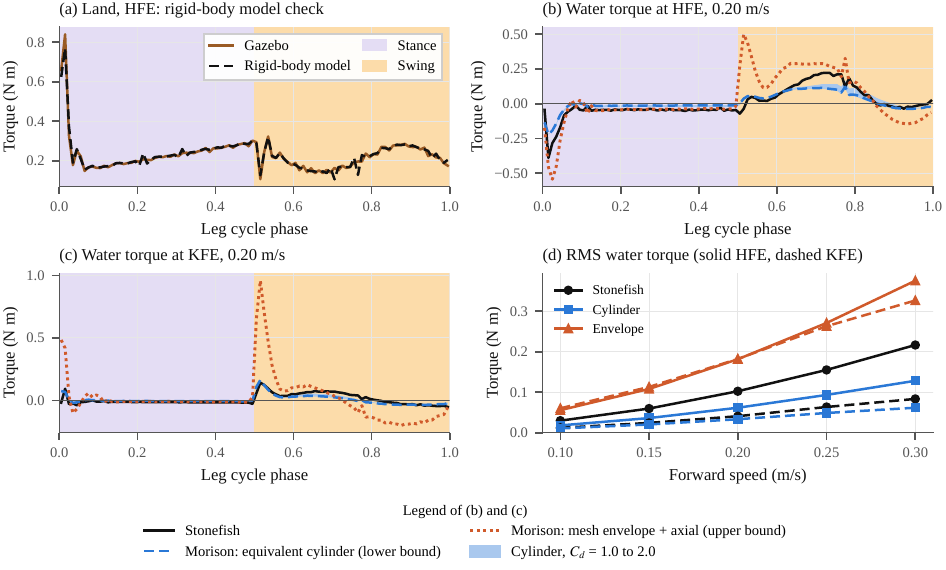}
  \caption{Analytic check: (a) rigid-body model against the Gazebo HFE torque on land; water torque at (b) HFE and (c) KFE at 0.20\,m/s, from Stonefish (Equation~\ref{eq:residual}) and from the two Morison estimates, the equivalent cylinder (lower bound) and the mesh envelope with axial flow (upper bound), with the legend below the figure; (d) RMS water torque against speed.}
  \label{fig:check}
\end{figure}

Because the motors feel the water through the joint torques, the check is made at the joints. First, a rigid-body inverse-dynamics model of the leg~\cite{featherstone2008} (link masses, inertias and gravity from the robot description, joint motion from the recording) was applied to the land run. It reproduces the Gazebo torques with correlation 0.98 (HFE) and 0.97 (KFE) and RMS differences of 0.016 and 0.010\,N\,m (Figure~\ref{fig:check}(a)), which validates the model and the kinematics. Adding the buoyancy of each link at its centre of volume, computed from the link mesh, gives 2.648 and 1.772\,N, identical to the Stonefish values. The water torque applied by Stonefish is then obtained as

\begin{equation}
  \tau_{w,j} = \tau_{rb+b,j} - \tau_{\mathrm{meas},j},
  \label{eq:residual}
\end{equation}

where $\tau_{rb+b}$ is the rigid-body torque including buoyancy and $\tau_{\mathrm{meas}}$ the torque recorded in water.

An independent, analytic estimate of the water torque is obtained with Morison's strip theory~\cite{morison1950}, in its form for a body moving in still water, where only an added-mass term and a drag term remain. Each link, i.e., the upper and lower link, is divided into 5\,mm strips along its axis. A strip moving with velocity $v_n$ and acceleration $a_n$ perpendicular to the link axis receives the force

\begin{equation}
  dF = -\left(\rho C_a A_a\,a_n + \tfrac12\rho C_d\,b\,|v_n|\,v_n\right)ds,
  \qquad C_a = 1,\; C_d = 1.2\;(Re_D \approx 2\times10^3\text{--}10^4),
  \label{eq:morison}
\end{equation}

where $A_a$ is the cross-sectional area that sets the added mass, $b$ the width of the strip facing the flow, $C_a$ the added-mass coefficient and $C_d$ the drag coefficient. The moments of these strip forces about the HFE and KFE axes give the analytic water torque.

The difficulty lies in the cross-section. The links are thin-walled shells, not solid cylinders: their material fills only 31--39\% of their outer outline. Since it is not known whether the water effectively sees the material or the outline, two section models are therefore used to obtain a lower and an upper estimate of the water torque~\cite{newman1977,hoerner1965}:

\begin{itemize}
  \item \textit{Lower bound (A):} Each link is replaced by a solid cylinder of
        the same volume (diameter 32\,mm for the thigh and 25\,mm for the
        shank).
  \item \textit{Upper bound (B):} Each strip takes the full outer outline of
        the link, approximated by an ellipse. Estimate B also includes motion
        along the link axis, which strip theory normally ignores: during
        stance the shank slides mostly along its own length (0.14\,m/s along
        the axis against 0.02--0.06\,m/s across it), and its 120\,mm wide
        flare pushes the water head-on.
\end{itemize}

Figures~\ref{fig:check}(b) and~\ref{fig:check}(c) compare the three water torques over one cycle at 0.20\,m/s, at the HFE and KFE joints. In both, the Stonefish torque (solid black) lies between the cylinder estimate (blue, dashed) and the envelope estimate (orange, dotted) and rises and falls at the same phases: it is small during stance, has its largest values at touch-down and lift-off, and follows the rise and fall of the swing. Its correlation with estimate A is 0.77--0.95 at HFE and 0.77--0.89 at KFE. 
Figure~\ref{fig:check}(d) extends the comparison to all speeds: from 0.10 to 0.30\,m/s the RMS Stonefish torque stays between the two bounds, at 1.6--1.7 times estimate A and 0.54--0.58 times estimate B at HFE, and 1.1--1.4 times A and 0.20--0.26 times B at KFE. Because these ratios hardly change with speed, the simulated water loads grow with speed as the theory predicts.

What the check cannot settle is the exact magnitude: the two bounds differ by a factor of three to five because of the thin-walled shape. The added mass shows this most clearly. For the shank, theory allows any value between 0.18\,kg (cylinder) and 0.78\,kg (envelope); Stonefish uses 0.13\,kg, below both, and its drag on the real link surface is correspondingly larger than in estimate A. How much this uncertainty affects the joint torques is examined next.

\subsection{Sensitivity to the drag coefficient and the added mass}

Section~5.1 leaves the added mass of the shank uncertain by a factor of several, and the drag coefficient is only an estimate. To see how much these two inputs matter, the water walk was repeated with one of them scaled and the other kept at its nominal value: the drag coefficient ($C_d = 1$ for cross-flow) and the added mass ($m_a$ = 0.201 and 0.133\,kg) were each multiplied by 0, 0.5, 1.5 and 2, at 0.20 and 0.30\,m/s. The factor 0 removes the effect completely, and the factor 2 for $C_d$ reaches the upper end of the band of Section~5.1. All three joints move in every run; Table~\ref{tab:sens} lists the change of the RMS torque of the two loaded joints, HFE and KFE.

The added mass is the parameter that matters. Removing it lowers the joint torques by about one third ($-30$\% to $-34$\%), and the torques change almost linearly with it, by about 17\% for every half of its nominal value. The drag coefficient matters much less: removing the drag lowers the torques by only 3 to 9\%, and doubling it raises them by 15 to 18\%. 
The percentages are nearly the same at both speeds, because the added-mass and drag forces grow with speed at similar rates (exponents 1.6 and 2.0, Section~4), so their share of the torque changes little.

A $\pm50$\% error in the added mass, therefore, changes the water joint torques by about $\pm17$\% and moves the HFE crossover speed of Section~4 from 0.23\,m/s to within about 0.21 to 0.27\,m/s. The conclusions of Sections~3 and~4 do not change, but the exact torque values depend mainly on the added mass, which is the parameter to measure first.

\begin{table}[htbp]
  \centering
  \caption{Change of the RMS joint torque in water when one hydrodynamic
  parameter is scaled (the other kept at its nominal value).}
  \label{tab:sens}
  \small
  \begin{tabular}{@{}llrrrr@{}}
    \toprule
    Scaled parameter & Joint, speed & $\times$0 & $\times$0.5 & $\times$1.5 & $\times$2 \\
    \midrule
    Drag coefficient $C_d$ & HFE, 0.20 m/s & $-8.0$\% & $-5.4$\% & $+8.0$\% & $+16.2$\% \\
                           & KFE, 0.20 m/s & $-4.3$\% & $+0.1$\% & $+10.3$\% & $+17.5$\% \\
                           & HFE, 0.30 m/s & $-8.5$\% & $-5.3$\% & $+7.0$\% & $+16.7$\% \\
                           & KFE, 0.30 m/s & $-3.0$\% & $-3.5$\% & $+6.6$\% & $+15.2$\% \\
    Added mass $m_a$       & HFE, 0.20 m/s & $-34.2$\% & $-17.2$\% & $+16.9$\% & $+33.2$\% \\
                           & KFE, 0.20 m/s & $-33.0$\% & $-17.5$\% & $+15.3$\% & $+31.5$\% \\
                           & HFE, 0.30 m/s & $-31.9$\% & $-17.2$\% & $+16.8$\% & $+34.5$\% \\
                           & KFE, 0.30 m/s & $-30.1$\% & $-16.0\%$ & $+16.3$\% & $+32.5$\%\\
    \bottomrule
  \end{tabular}
\end{table}

\section{Discussion and Conclusions}

One leg of an amphibious quadruped was simulated on land and in water using the same robot model, leg trajectory and controller, so that differences in the joint loads arise solely from the medium. Because the leg is almost neutrally buoyant, water removes the gravity load that dominates on land: at 0.20\,m/s the RMS KFE torque falls from 0.26 to 0.08\,N\,m and the HAA torque from 0.26 to 0.04\,N\,m, with the same joint and foot motion. The water loads grow with speed, however, and the RMS hip torque and the joint work in water exceed their land values above 0.23 and 0.25\,m/s, respectively. The added-mass force is the largest water force, and its peaks come from the velocity steps of the leg trajectory. An analytic Morison check places the simulated water torques between lower and upper bounds, with the expected shape and speed scaling, and a sensitivity study shows that the added mass controls the result, not the drag coefficient. These findings have three consequences for the design of an amphibious leg.

\textit{Actuator sizing:} The peak torque is set by land, where the leg is accelerated against gravity, at every speed tested. The continuous (RMS) torque, which sets motor heating, is set by land at slow speeds but by water above about 0.23\,m/s, and it grows steeply from there. Actuators sized for land walking are therefore adequate in water up to this speed; for faster underwater walking their continuous rating, not their peak rating, must be increased.

\textit{Walking strategy:} Because the water loads grow with the stepping frequency, slow walking with long strides is cheaper underwater, whereas on land the cost per cycle hardly depends on speed. A controller that adapts the gait to the medium could exploit this.

\textit{Gait shaping:} A large part of the added-mass load comes from the abrupt velocity changes at lift-off and touch-down. A swing trajectory whose velocity matches the stance velocity at both ends should reduce these peaks. Because the added-mass force exceeds the drag force at all tested speeds, this is the most direct way to reduce the water loads.

\section{Future Work}
The results of this study will guide the hardware validation of the single leg, in which its joint torques will be measured on land and in a water tank and compared with the simulated values. These tests will also measure the added mass of the leg links, the parameter that controls the water loads (Section~5.2), and will be used to calibrate the hydrodynamic model. The calibrated simulation will then be extended to a new design of the full amphibious quadruped, including body motion and ground contact on land and on the seabed. Finally, the quadruped will be built and verified on land and in water, using swing trajectories designed to reduce the added-mass loads.

\section*{Acknowledgments}
Generative AI technology was used in the preparation of this document to improve grammar and refine text. All authors retain full responsibility for the content and its presentation.

{\small
\bibliographystyle{IEEEtran}   
\bibliography{reference}       

@article{picardi2020,
  author  = {Picardi, Giacomo and Chellapurath, Mrudul and Iacoponi, Saverio and Stefanni, Sergio and Laschi, Cecilia and Calisti, Marcello},
  title   = {Bioinspired underwater legged robot for seabed exploration with low environmental disturbance},
  journal = {Science Robotics},
  volume  = {5},
  number  = {42},
  pages   = {eaaz1012},
  year    = {2020}
}

@inproceedings{jun2013,
  author    = {Jun, Bong-Huan and Shim, Hyungwon and Kim, Banghyun and Park, Jin-Yeong and Baek, Hyuk and Yoo, Seong and Lee, Pan-Mook},
  title     = {Development of seabed walking robot {CR200}},
  booktitle = {Proc. MTS/IEEE OCEANS},
  address   = {Bergen, Norway},
  year      = {2013}
}

@article{dudek2007,
  author  = {Dudek, Gregory and others},
  title   = {{AQUA}: An amphibious autonomous robot},
  journal = {Computer},
  volume  = {40},
  number  = {1},
  pages   = {46--53},
  year    = {2007}
}

@article{calisti2017,
  author  = {Calisti, Marcello and Corucci, Francesco and Arienti, Andrea and Laschi, Cecilia},
  title   = {Dynamics of underwater legged locomotion: modeling and experiments},
  journal = {Bioinspiration \& Biomimetics},
  volume  = {12},
  number  = {1},
  pages   = {016003},
  year    = {2017}
}

@inproceedings{koenig2004,
  author    = {Koenig, Nathan and Howard, Andrew},
  title     = {Design and use paradigms for {Gazebo}, an open-source multi-robot simulator},
  booktitle = {Proc. IEEE/RSJ Int. Conf. Intelligent Robots and Systems (IROS)},
  pages     = {2149--2154},
  year      = {2004}
}

@inproceedings{cieslak2019,
  author    = {Cie{\'s}lak, Patryk},
  title     = {Stonefish: An advanced open-source simulation tool designed for marine robotics, with a {ROS} interface},
  booktitle = {Proc. OCEANS 2019},
  address   = {Marseille, France},
  year      = {2019}
}

@inproceedings{manhaes2016,
  author    = {Manh{\~a}es, Musa Morena Marcusso and Scherer, Sebastian A. and Voss, Martin and Douat, Luiz Ricardo and Rauschenbach, Thomas},
  title     = {{UUV} {Simulator}: A {Gazebo}-based package for underwater intervention and multi-robot simulation},
  booktitle = {Proc. OCEANS 2016 MTS/IEEE},
  address   = {Monterey, CA, USA},
  year      = {2016}
}

@article{chitta2017,
  author  = {Chitta, Sachin and others},
  title   = {ros\_control: A generic and simple control framework for {ROS}},
  journal = {Journal of Open Source Software},
  volume  = {2},
  number  = {20},
  pages   = {456},
  year    = {2017}
}

@article{macenski2022,
  author  = {Macenski, Steven and Foote, Tully and Gerkey, Brian and Lalancette, Chris and Woodall, William},
  title   = {Robot {Operating} {System} 2: Design, architecture, and uses in the wild},
  journal = {Science Robotics},
  volume  = {7},
  number  = {66},
  pages   = {eabm6074},
  year    = {2022}
}

@book{schlichting2017,
  author    = {Schlichting, Hermann and Gersten, Klaus},
  title     = {Boundary-Layer Theory},
  edition   = {9th},
  publisher = {Springer},
  address   = {Berlin},
  year      = {2017}
}

@book{featherstone2008,
  author    = {Featherstone, Roy},
  title     = {Rigid Body Dynamics Algorithms},
  publisher = {Springer},
  address   = {New York},
  year      = {2008}
}

@article{morison1950,
  author  = {Morison, J. R. and O'Brien, M. P. and Johnson, J. W. and Schaaf, S. A.},
  title   = {The force exerted by surface waves on piles},
  journal = {Petroleum Transactions, AIME},
  volume  = {189},
  pages   = {149--154},
  year    = {1950}
}

@book{newman1977,
  author    = {Newman, John Nicholas},
  title     = {Marine Hydrodynamics},
  publisher = {MIT Press},
  address   = {Cambridge, MA},
  year      = {1977}
}

@book{hoerner1965,
  author    = {Hoerner, Sighard F.},
  title     = {Fluid-Dynamic Drag},
  publisher = {Published by the author},
  year      = {1965}
}

@inproceedings{hutter2016,
  author    = {Hutter, Marco and others},
  title     = {{ANYmal} -- a highly mobile and dynamic quadrupedal robot},
  booktitle = {Proc. IEEE/RSJ Int. Conf. Intelligent Robots and Systems (IROS)},
  pages     = {38--44},
  year      = {2016}
}

@inproceedings{bledt2018,
  author    = {Bledt, Gerardo and Powell, Matthew J. and Katz, Benjamin and Di Carlo, Jared and Wensing, Patrick M. and Kim, Sangbae},
  title     = {{MIT} {Cheetah} 3: Design and control of a robust, dynamic quadruped robot},
  booktitle = {Proc. IEEE/RSJ Int. Conf. Intelligent Robots and Systems (IROS)},
  pages     = {2245--2252},
  year      = {2018}
}

@book{murphy2014,
  author    = {Murphy, Robin R.},
  title     = {Disaster Robotics},
  publisher = {MIT Press},
  address   = {Cambridge, MA},
  year      = {2014}
}

@article{ijspeert2007,
  author  = {Ijspeert, Auke Jan and Crespi, Alessandro and Ryczko, Dimitri and Cabelguen, Jean-Marie},
  title   = {From swimming to walking with a salamander robot driven by a spinal cord model},
  journal = {Science},
  volume  = {315},
  number  = {5817},
  pages   = {1416--1420},
  year    = {2007}
}

@article{baines2022,
  author  = {Baines, Robert and Patiballa, Sree Kalyan and Booth, Joran and Ramirez, Luis and Sipple, Thomas and Garcia, Andonny and Fish, Frank and Kramer-Bottiglio, Rebecca},
  title   = {Multi-environment robotic transitions through adaptive morphogenesis},
  journal = {Nature},
  volume  = {610},
  pages   = {283--289},
  year    = {2022}
}
}

\end{document}